\documentclass{article}
\usepackage[T1]{fontenc}
\usepackage[preprint]{neurips_2026}
\usepackage{amsmath,amssymb,amsthm,booktabs,graphicx,microtype,float,times}
\usepackage{enumitem}
\usepackage{hyperref}
\usepackage{url}
\usepackage{xcolor}
\usepackage[ruled,vlined]{algorithm2e}

\ifdefined\theorem
\else
  \newtheorem{theorem}{Theorem}
\fi

\hypersetup{
  colorlinks=true,
  linkcolor=blue!55!black,
  citecolor=blue!55!black,
  urlcolor=blue!55!black
}

\title{Sequential Pretraining Favors Large Models}

\author{%
  Mohnish Harwani \\
  Purdue University \\
  \texttt{mharwan@purdue.edu}
  \And
  Yujia Zheng \\
  University of Illinois Urbana--Champaign \\
  \texttt{yujiaz@illinois.edu}
}

\newcommand{\Dgood}{\mathcal{D}_{\mathrm{g}}}

\begin{document}
\maketitle
 
\begin{abstract}
Large neural networks often acquire capabilities that small models fail to learn. Does this stem from large models learning more representative features, or from being more robust to unaccounted-for adverse effects introduced during training? We define and quantify one such adverse effect, \textit{primacy bias}, as the extent to which exposure to early data distributions impairs later learning.
We show that small models can allocate learning capacity inefficiently toward early distributions, whereas sufficiently overparameterized models are robust to this effect. This inefficiency is particularly consequential in pretraining, where foundation models often encounter heterogeneous data distributions sequentially rather than jointly. As a result, small foundation models can struggle to learn distributions encountered late in training, which is particularly harmful when later data emphasizes desirable capabilities such as code, mathematics, and reasoning. Motivated by these findings, we introduce Exposure Therapy (ET), a simple regularization that promotes more efficient allocation of learning capacity during sequential pretraining. We demonstrate that ET improves foundation models' performance on late data distributions as well as overall capability in models up to the billion-parameter scale. Overall, our results suggest that some benefits of large foundation models may arise from greater robustness to adverse training effects, rather than from learning more representative features, and that improved training algorithms can recover some of these advantages in smaller models.
\end{abstract}

\section{Introduction}
\label{sec:introduction}
Large machine learning models often learn behaviors that smaller models fail to acquire, motivating the continued scaling of models despite substantial training and inference costs \citep{wei2022emergent, huang2026larger}. However, scale may provide more than additional representational power: it can also make learning more robust such that it can tolerate a broader set of training algorithms through which data are presented, converging to good solutions more predictably \citep{lourie2026smallscale,wen2026fantastic}. 
One such source of variation in training algorithms is data order. While existing work in curriculum learning has demonstrated how data order can alter generalization \citep{elman1993starting,bengio2009curriculum,hacohen2019curriculum,wu2021curricula,jia2026curriculum,finzi2026epiplexity}, it treats data order as a design choice, whereas pretraining algorithms often enforce it as a constraint.

Loss of plasticity provides an explanation for the variation in performance that data order can introduce. Several lines of work have demonstrated that neural networks gradually lose plasticity as training progresses, motivating representative data to be seen early \citep{dohare2024plasticity,lyle2023plasticity}. 
Other results suggest that capacity may influence this loss of plasticity. Neural networks can preferentially learn simple or high-utility structure early, and gradients associated with dominant features can suppress competing signals \citep{arpit2017memorization,kalimeris2019complexity,refinetti2023complexity,pezeshki2021gradient,shah2020simplicity}. Overparameterization can improve tolerance to noise and reduce interference or forgetting across tasks \citep{nakkiran2020double,mirzadeh2022wide,ramasesh2022scale,mohri2026bitter,marek2026forgetting}. Most recently, \citet{huang2026larger} demonstrated that large model capacity can provide resistance to overwriting useful learned features. While nearly all of these works study data order in terms of retaining previously learned information, they do not address the combined relationship between data order and model capacity in learning new information.

\begin{figure}[t]
    \centering
    \includegraphics[width=\linewidth]{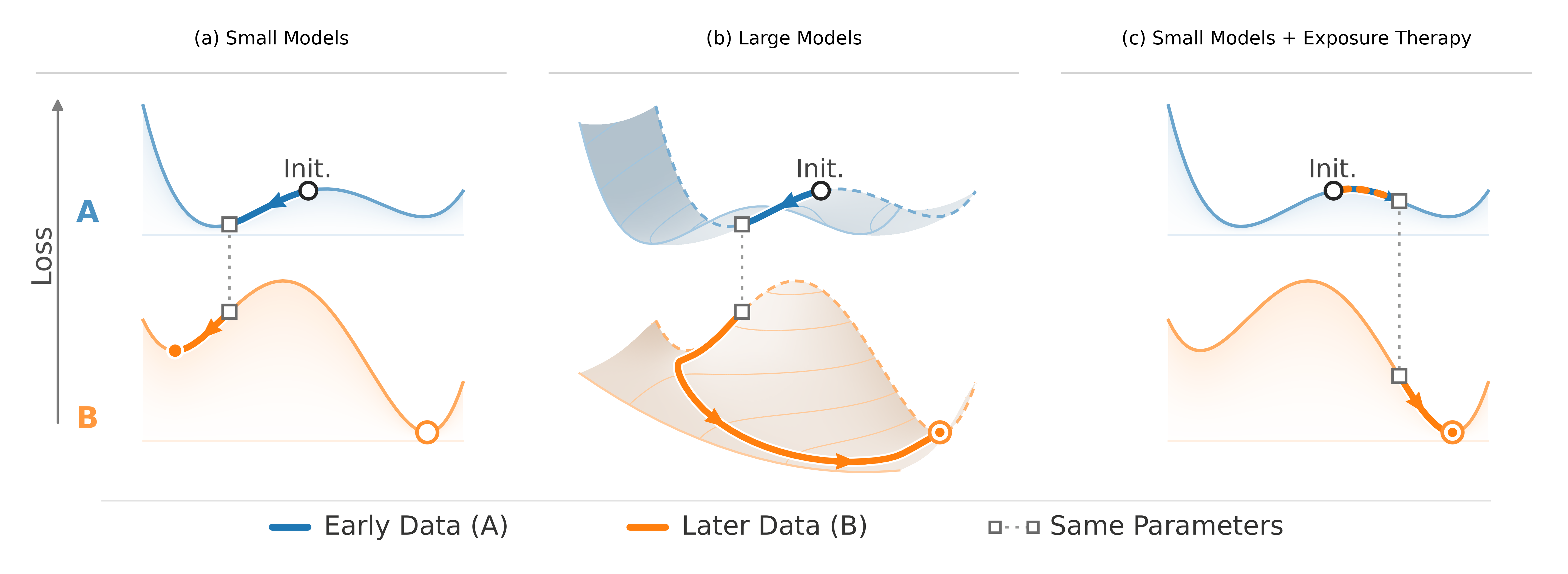}
    \caption{
        \textbf{Conceptual illustration of primacy bias and Exposure Therapy.}
        Small models may become constrained by parameters recruited for early data, making later distributions harder to fit. Additional capacity can allow optimization to lead to alternative solutions, while Exposure Therapy encourages small models toward parameters that remain compatible with later learning.
    }
    \label{fig:conceptual-overview}
\end{figure}

In this work, we define and quantify \textit{primacy bias}: the extent to which early data distributions can impair a model’s ability to fit distributions encountered later in training. We study the role of scaling in primacy bias, and demonstrate that overparameterized models are far more robust to primacy bias while small models are not. These results suggest that some benefits of large models may arise from greater robustness to biases introduced in training, as opposed to learning more representative solutions. In foundation model pretraining, the effects of primacy bias are further amplified.
While training foundation models often utilizes several heterogeneous data sources \citep{xie2023doremi,feng2024twophase,ibrahim2024continual}, pretraining often constrains this by organizing the data into a temporal sequence, where foundation models encounter heterogeneous data distributions sequentially rather than jointly. Models may be exposed to different subsets of the training distribution in distinct stages, where early training may emphasize broad natural-language data, and later stages increasingly incorporate more specialized domains such as code, mathematics, tool use, and agentic behavior. We find that this sequential training constraint favors large and overparameterized foundation models in terms of learning in late-stage data. Specifically, we observe that when small models lack the capacity to fit all data they encounter during training, learning new distributions increasingly requires repurposing neurons specialized to earlier data. This process of repurposing neurons is hindered by loss of plasticity and results in small foundation models fitting late-stage data less effectively.

To address these failures, we introduce \emph{Exposure Therapy} (ET), which sparsely replaces training updates with a sample of downstream data distributions that models would otherwise encounter later in training. An important distinction is that Exposure Therapy does not remove this data from later training where it might otherwise be seen, and simply re-uses existing data. ET encourages models to learn structure specific to downstream data distributions during its critical learning period, reserving capacity for later acquisition. Unlike replay methods \citep{lopezpaz2017gem,chaudhry2019tiny,chaudhry2021hal}, ET focuses on improving performance related to newly encountered distributions, as opposed to preserving previously acquired capabilities, and closing capacity-dependent performance gaps in pretraining. We show that Exposure Therapy improves learning new data distributions as well as overall performance in foundation models, allowing models to recover part of the robustness otherwise obtained through additional parameters, with negligible extra computational cost. Figure~\ref{fig:conceptual-overview} visualizes these effects. We demonstrate that Exposure Therapy improves pretraining in foundation models up to the billion-parameter scale.

\section{Preliminaries}
\label{sec:preliminaries}

We begin with our problem setup. Our goal is to quantify the extent to which early training distributions can impair a model's fit on later training distributions, and understand how that impairment changes across model scale. We consider two training conditions. In \emph{target-first} training, a model is trained on the target distribution during Phase~1; in \emph{OOD-first} training, it is instead trained on an out-of-distribution (OOD) distribution, which we use to refer to any distribution distinct from the target distribution being evaluated. After Phase~1, both conditions follow the same target-dominated Phase~2 training trajectory, without resetting model parameters, optimizer state, or the learning-rate schedule. For a model with \(N\) parameters, let \(\theta_{\mathrm{tgt}}^{(N)}(t)\) and \(\theta_{\mathrm{ood}}^{(N)}(t)\) denote its parameters after \(t\) training steps under target-first and OOD-first training, respectively. Given target loss \(\ell\), we define \emph{primacy sensitivity} as the persistent target-performance gap
\begin{equation}
    \Delta_{\mathrm{prim}}(N)
    :=
    \lim_{t\rightarrow\infty}
    \left[
        \ell\!\left(\theta_{\mathrm{ood}}^{(N)}(t)\right)
        -
        \ell\!\left(\theta_{\mathrm{tgt}}^{(N)}(t)\right)
    \right].
    \label{eq:primacy-sensitivity}
\end{equation}
A positive \(\Delta_{\mathrm{prim}}(N)\) therefore indicates that exposure to the OOD distribution during early training persistently impairs the model's fit on the target distribution. For classification experiments, \(\ell\) denotes classification error, while for language modeling it denotes held-out cross-entropy. Empirically, we approximate the limit in Equation~\eqref{eq:primacy-sensitivity} by training the shared Phase~2 trajectory until target performance has converged.

Our controlled experiments in Section~\ref{sec:controlled} illustrate how primacy sensitivity changes across scale by approximating this behavior in MLPs. Section~\ref{sec:foundation} extends these results to pretraining settings, while Section~\ref{sec:anchor} introduces Exposure Therapy in order to provide robustness to pretraining without increasing the number of parameters.

\section{The Data Order-Capacity Relationship}
\label{sec:controlled}

Primacy bias is not specific to the scale of foundation models or pretraining. More generally, it arises when early exposure to one distribution changes how a model can fit another distribution. We first isolate this effect in vision tasks over MLPs, illustrating how primacy bias is a general property of neural networks, and how circumventing capacity used to fit early distributions allows large models to become more robust to this effect. These task setups are meant to make primacy bias easier to analyze mechanistically and understand why it can be scale dependent, rather than replicate the dynamics of pretraining. Section~\ref{sec:foundation} then extends these basic effects to foundation models, where primacy bias can arise from sequential data order constraints in pretraining and result in poor allocation of learning capacity over pretraining data.

\subsection{Experimental Setup: Isolating Training-History Sensitivity}
\label{sec:protocol}

We study this relationship in controlled vision experiments on MNIST, Fashion-MNIST, and KMNIST \citep{lecun1998mnist,xiao2017fashion,clanuwat2018kmnist},
varying MLP width while holding the remaining training procedure fixed. For these experiments, the OOD distribution consists of real inputs paired
with fixed random labels, creating substantial early learning with little structure transferable to the target task. During Phase~2, every fourth update is drawn from a separately sampled OOD set and the remaining updates
from the target distribution, providing a weak persistent signal for the earlier structure. We train Phase~2 substantially longer than Phase~1 and
evaluate models after target performance has converged. Full experimental details are provided in Appendix~\ref{app:toy_setup}.

\subsection{Primacy bias decreases with model scale}
\label{sec:collapse}

\begin{figure}[h]
    \centering
    \includegraphics[width=\linewidth]{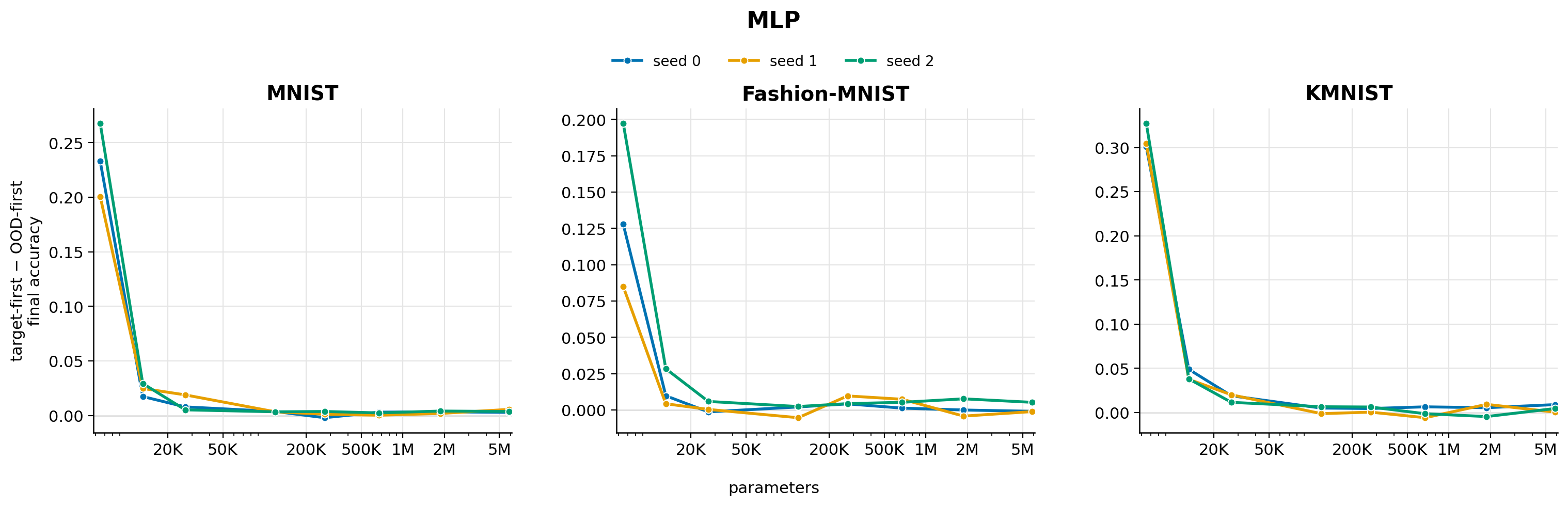}
    \caption{
        \textbf{Sensitivity to harmful early data decreases with model scale.} Primacy sensitivity \(\Delta_{\mathrm{prim}}\) compares final target performance under target-first and OOD-first training. Positive values indicate that the early OOD phase impaired later target learning. Columns indicate different datasets; curves within each panel correspond to different seeds. Small models exhibit significant primacy sensitivity, while the gap attenuates sharply with increasing capacity.
    }
    \label{fig:junkdiff}
\end{figure}

The results of these experiments demonstrate how primacy bias falls sharply with model scale (Figure~\ref{fig:junkdiff}). Small OOD-first models perform substantially worse in terms of accuracy on later distributions compared to target-first models of the same size. As model capacity increases, this gap sharply decreases. At sufficient levels of overparameterization, final performance becomes nearly insensitive to which distribution was encountered first.

Interestingly, in these settings, small models have enough representational capacity to learn late distributions well. However, they fail to adapt to a new distribution of data while sufficiently overparameterized models can. Overparameterized models are more robust to OOD data seen in early training, and are able to learn new distributions significantly better.

Importantly, this difference is not explained by a difference in plasticity in large models. We find that models of all scales tend to exhibit low accuracy on OOD data seen early (Appendix~\ref{app:full-corruption-retention}), suggesting the existence of a latent effect on subsequent learning in small models, even after the behavior learned during that phase is no longer expressed.

This result illustrates a benefit of overparameterization beyond representational capacity alone. A smaller model may contain enough parameters to fit the target distribution in isolation, yet still fail to reach the same solution when part of its training capacity was strongly recruited by an earlier distribution. Larger models are substantially more robust to this dependence on training history.

\subsection{Representational overlap decreases with model scale}
\label{sec:mechanismresults}

To understand how overparameterized models can become robust to primacy bias, we analyze hidden-unit reuse in MLPs. Specifically, at different thresholds, we rank units by post-activation magnitude on the OOD distribution at the end of Phase-1 and on the target distribution at the end of Phase-2, then measure overlap among the top 1\%, 5\%, and 10\% most active units. This allows us to understand how models are forced to reuse existing parameters to map new distributions, and where any latent effects might lie. Full definitions and experimental details are provided in Appendix~\ref{sec:overlap}.

\begin{figure}[h]
    \centering
    \includegraphics[width=\linewidth]{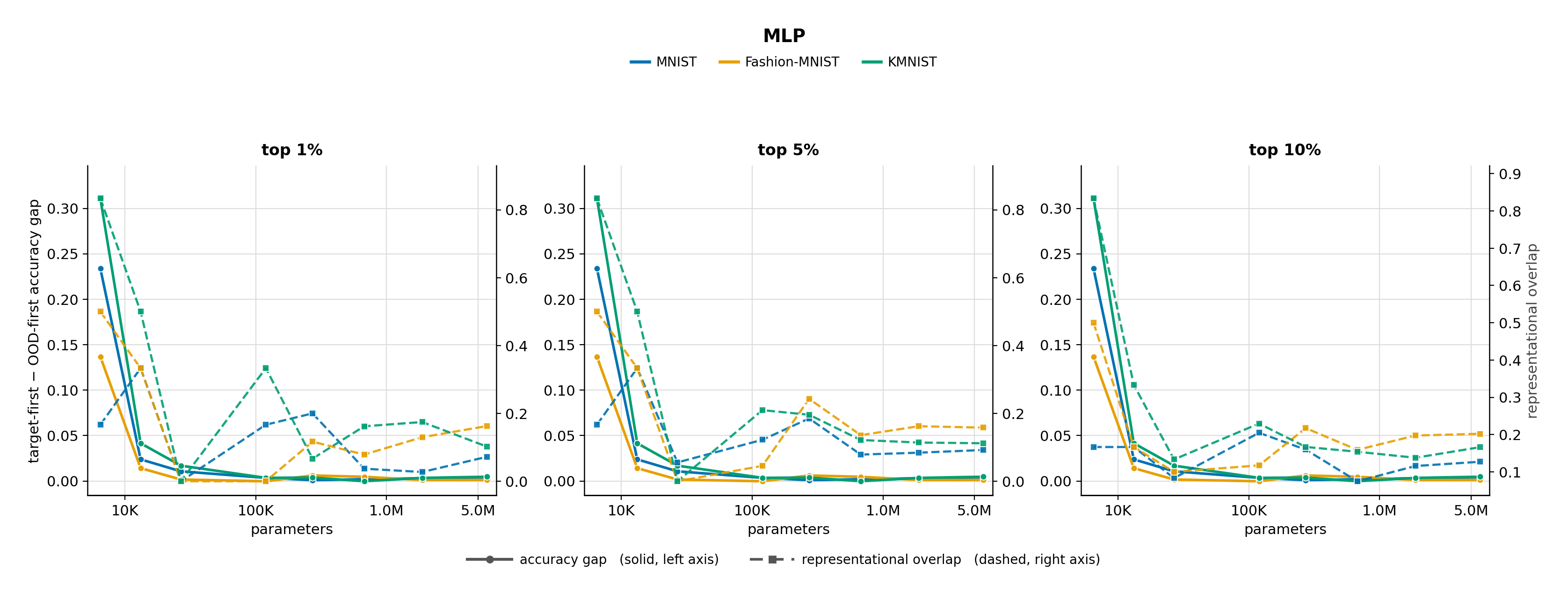}
    \caption{
        \textbf{Large models recruit a less-overlapping route for downstream learning.}
        Solid curves with circular markers indicate primacy sensitivity \(\Delta_{\mathrm{prim}}\), using the left axis; dashed curves with square
        markers show cross-phase representational overlap on the right axis. The reduction in sensitivity roughly aligns with the reduction in overlap
        between the units recruited during early OOD training and those recruited after subsequent target training. This correspondence supports
        the hypothesis that additional capacity allows larger models to route downstream learning through a less-overlapping set of units, rather than repurposing the units most strongly shaped by the early OOD distribution. At high activation thresholds, this signal becomes less accurate and rises in larger models. 
    }
    \label{fig:mechanism}
\end{figure}

Representational overlap in post-activation neurons generally decreases with primacy bias changes across model scale (Figure~\ref{fig:mechanism}). Small models exhibit both the strongest downstream impairment and the greatest reuse of units recruited during Phase-1. Both quantities decline as scale increases. This pattern is qualitatively consistent across activation thresholds.

More broadly, these results show that the acquisition of representative features in late distributions doesn't just depend on whether the model has sufficient capacity in principle, but also on how much of that capacity is \textit{adaptable} after early learning. Overparameterized models can largely circumvent capacity that fit early training distributions, and use other capacity they have available. Small models must often repurpose existing capacity to fit a later distribution, which can be difficult in late training due to factors like loss of plasticity.

\section{Primacy Bias in Foundation Model Pretraining}
\label{sec:foundation}

Section~\ref{sec:controlled} establishes a general sensitivity of primacy bias across model scale, demonstrating how capacity-constrained models are more vulnerable to this effect. Under foundation model pretraining, we demonstrate how this becomes particularly consequential. Pretraining corpora often contain several heterogeneous domains that are trained on sequentially, rather than jointly. For example, while early learning might emphasize natural language, later learning may emphasize desirable qualities such as code, mathematics, tool-use, and reasoning. A distribution introduced late must therefore be learned after earlier data have already shaped the model during its most plastic period. We show that, from 100M to 1B parameters, substantial early training on another useful distribution can impair how capacity-constrained models later fit another distribution.

\subsection{Sequential-Domain Pretraining Protocol}
\label{sec:foundation-protocol}

We extend the two-phase setting of Section~\ref{sec:controlled} to causal language-model pretraining at 100M, 500M, and 1B parameters. For each model scale, the first half of training consists either of English data (\emph{English-first}) or one of four OOD corpora (\emph{OOD-first}): code, mathematics, German, or Finnish. All models then train on the same majority English distribution for the second half of training. Each phase of training sees an equal training budget which we observe is largely sufficient for convergence. We evaluate how well models fit an English data distribution under both of these conditions where only the early data distribution varies.

Furthermore, we increase both the amount of training data and the optimization budget with model size. Thus, the 1B model is not simply a more overparameterized model trained on the same problem as the 100M model; it is trained on a correspondingly larger corpus. These experiments test whether models that remain capacity-constrained relative to their training data continue to exhibit sensitivity to the distributions encountered early in pretraining. Full experimental details are located in Appendix~\ref{sec:pretraining-details}.

\begin{figure}[h]
    \centering
    \includegraphics[width=\linewidth]{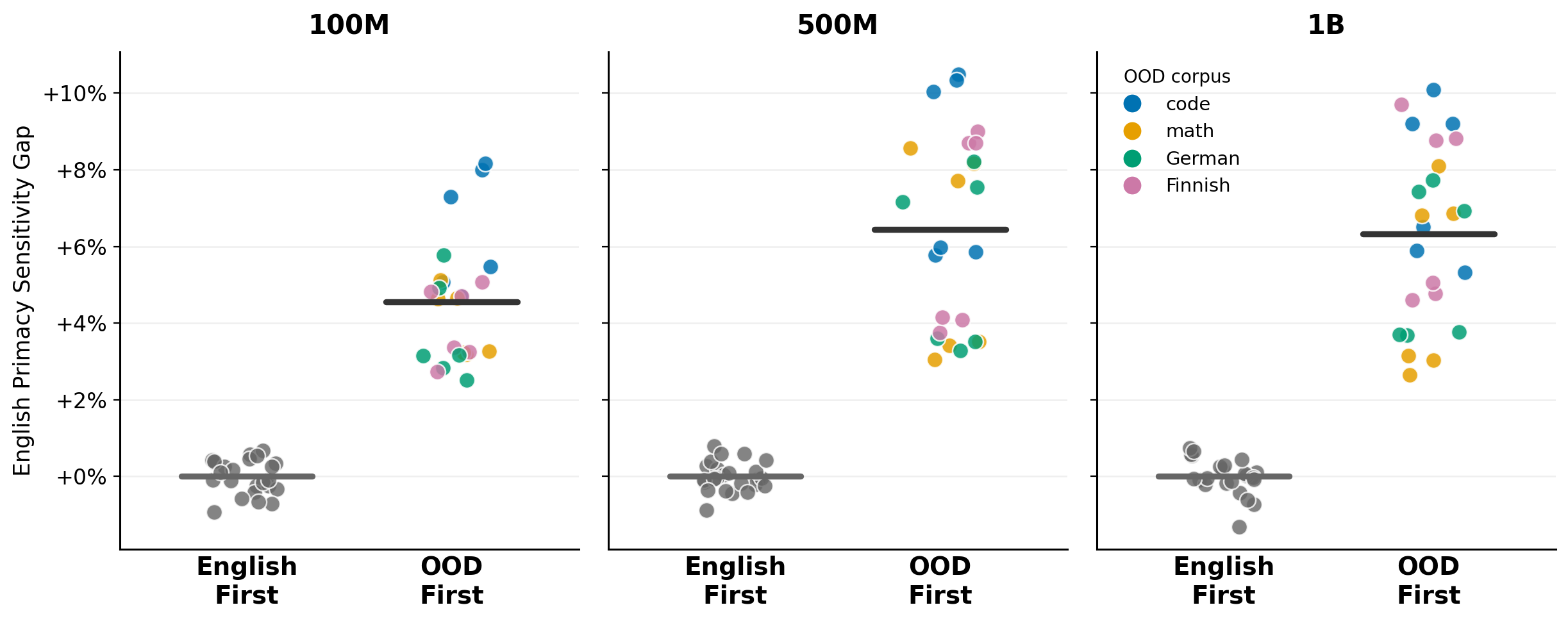}
    \caption{
        \textbf{Capacity-constrained foundation models remain sensitive to data encountered early in pretraining.}
        The English primacy sensitivity gap reports the percentage increase in final English loss relative to the mean English-first loss at the same model scale.
        During the first half of training, models see either English or an OOD corpus consisting of code, mathematics, German, or Finnish; all models then train on a majority English distribution during the second half. Points denote individual runs, colors indicate the OOD corpus, and horizontal bars denote group means. Across 100M, 500M, and 1B parameter models, OOD-first pretraining produces substantially higher final English loss.
        Training data and optimization budget increase with model size, keeping models capacity-constrained relative to their respective training problems.
    }
    \label{fig:foundation-primacy}
\end{figure}

Figure~\ref{fig:foundation-primacy} shows how training on OOD data first in pretraining consistently impairs the acquisition of the English distribution encountered later. At all three parameter scales, models that spend the first half of training on another distribution exhibit substantially higher final English loss. The magnitude of this impairment varies across OOD corpora. Overall, these results demonstrate how primacy sensitivity in neural networks can introduce a practical challenge when paired with temporal pretraining constraints in training less overparameterized foundation models. Useful early data can make capacity-constrained models substantially worse at fitting distributions encountered later, which is especially harmful as later distributions often better match tasks that have real-world applications. Because removing useful early distributions is not a viable solution, an effective pretraining algorithm should instead allow models to learn them while preserving capacity for distributions that will receive greater emphasis later, motivating \textit{Exposure Therapy}.

\section{Exposure Therapy}
\label{sec:anchor}

The preceding sections demonstrate how primacy bias in neural networks can impair less overparameterized foundation models fit in late training distributions in favor of earlier ones. This phenomenon motivates a direct intervention: expose the model to a representative sample of a later training distribution while its representation is still highly plastic, allowing some capacity to become reserved for that distribution before early learning dominates the representation that a model ultimately develops.

We introduce \emph{Exposure Therapy} (ET), a data-level regularization that intermittently replaces ordinary training updates with examples drawn from a distribution that would otherwise receive substantial training later. At each optimization step, the ordinary training batch is replaced by a batch from the \textit{exposure set} with probability \(a\), which we call the \emph{exposure rate}. Exposure updates replace rather than supplement ordinary updates, so ET does not increase the number of optimization steps. The data used in the exposure set remain in the late data corpus to be encountered again during training, allowing ET to influence optimization without requiring additional training data.

\begin{figure}[t]
    \centering
    \includegraphics[width=\linewidth]{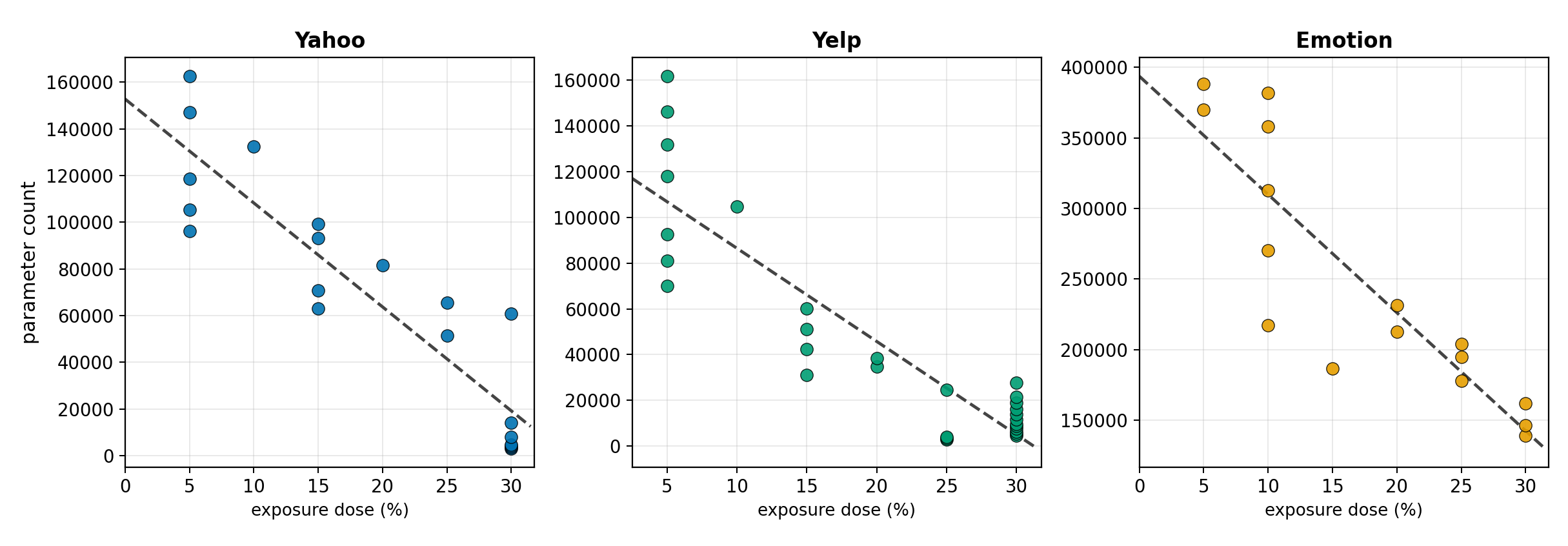}
    \caption{\textbf{Less overparameterized models require stronger Exposure Therapy.} Each point pairs a model's parameter count with the smallest tested exposure dose required to recover \(90\%\) of the corresponding target-first model's performance. We evaluate Transformer classifiers on Yahoo Answers, Yelp, and Emotion. Dashed lines denote dataset-specific linear fits. Across all three datasets, the required exposure dose increases as model capacity decreases.}
    \label{fig:anchor-threshold}
\end{figure}

Although ET is motivated with the goal of learning better allocations of structure across data distributions during the early critical learning period, we apply exposure throughout the training trajectory. While pretraining can make phase boundaries between different data boundaries clear, the precise duration of high-plasticity periods during which capacity allocation is most sensitive is still generally unknown. Maintaining sparse exposure avoids requiring an explicit estimate of when this period ends. Full pseudocode and analysis of these controls are provided in Appendix~\ref{app:expose_details}.

Figure~\ref{fig:anchor-threshold} demonstrates the relationship between model capacity and the amount of ET required on simple language tasks. Models with substantial capacity relative to the target distribution either recover from harmful early learning without ET or require only minimal exposure. As capacity decreases, the exposure required to recover the same fraction of performance increases, reaching \(20\%\)--\(30\%\) in the most capacity-constrained settings. This inverse relationship is consistent across all three language tasks. Full experimental details are in Appendix~\ref{sec:et_scale}.

\subsection{Exposure Therapy in Foundation Model Pretraining}
\label{sec:foundation-results}

We apply ET to the sequential-domain pretraining setting of Section~\ref{sec:foundation-protocol}. The exposure set consists of English data that would ordinarily be encountered during the later phase of training. We fix the exposure set to \(10\%\) of the target data training corpus and the exposure rate of \(30\%\). Exposure updates remain active throughout both phases.  We evaluate ET at the billion-parameter scale across code, mathematics, German, and Finnish OOD corpora and the FineWeb and TinyStories English target distributions. We further extend these results and evaluate this phenomenon at the 100M and 500M parameter scales in Appendix~\ref{app:more_scales}.

\begin{figure}[h]
    \centering
    \includegraphics[width=\linewidth]{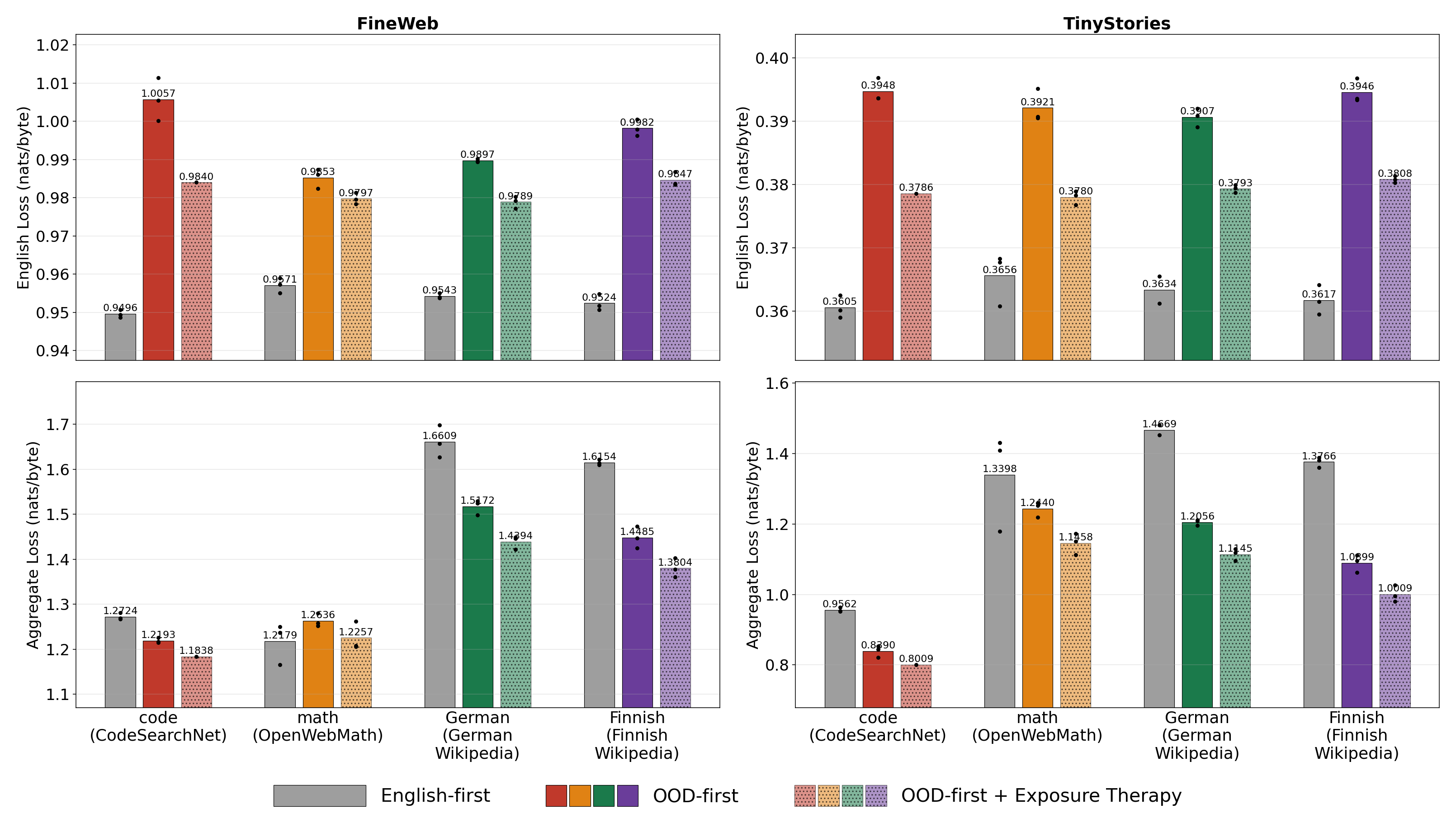}
    \caption{\textbf{Exposure Therapy improves both later-distribution and aggregate performance in 1B parameter models.} Top: held-out English loss after sequential pretraining on FineWeb and TinyStories; lower is better. Bottom: aggregate loss, computed across the Phase-1 OOD corpus and the English target distribution; lower is better. Within each OOD-corpus group, bars show English-first, OOD-first, and OOD-first training with a \(30\%\) Exposure Therapy rate. Black points denote individual runs. OOD-first training consistently increases subsequent English loss, while ET reduces this impairment across every tested OOD and target distribution. At the same time, OOD-first training generally benefits aggregate performance by learning both useful distributions, and ET further reduces aggregate loss across all tested conditions. ET therefore improves acquisition of the later distribution without sacrificing the useful structure learned earlier.}
    \label{fig:foundation-et}
\end{figure}

Figure~\ref{fig:foundation-et} evaluates both subsequent English learning and aggregate performance at the 1B parameter scale. Across every OOD corpus and both English target distributions, OOD-first training produces higher English loss than the corresponding English-first reference, while ET consistently reduces this impairment. Importantly, these gains do not come from simply shifting capacity away from the initial OOD distribution. Exposure Therapy also 
results in better overall test performance over all training distributions. Rather than trading performance between the two stages, ET generally produces a more efficient allocation of learning capacity that improves fit over late data as well as overall test performance in pretraining. These benefits come at negligible extra computational cost and require no additional data.

\section{Limitations}

Our mechanistic evidence demonstrates that parameters that fit early distributions are often avoided in sufficiently overparameterized models, providing a possible explanation for why large models are more robust to primacy bias. However, these results are limited in scope to MLPs where feature allocation is better understood. Furthermore, understanding the precise effect as to why these parameters are avoided or difficult to repurpose remains future work. Our pretraining experiments also use only two data distributions, whereas realistic pretraining may involve many sequential stages over many more distributions. Due to computational cost, we evaluate Exposure Therapy with fixed hyperparameters in foundation models and do not rigorously identify how to tune hyperparameters in pretraining. Instead, we use toy language models to study how suitable exposure strength varies with scale, and while exact hyperparameters might not transfer, we predict that their qualitative dependence on model scale will persist.
Our pretraining setup further assumes that Phase~2 retains a small fraction of OOD data or otherwise some weak training signal to retain earlier structure, and that sampling is shuffled within rather than across distributions, and that useful OOD data precedes English, whereas frontier pretraining often uses the opposite ordering. We perform ablation experiments without these assumptions to more closely match practical pretraining, demonstrating that ET improves pretraining in these settings as well, although in some cases to a reduced degree. While this still might not extend to all forms of pretraining, it demonstrates the broader applicability of this work under other pretraining pipelines.

\section{Conclusion}
\label{sec:conclusion}

Large models often acquire capabilities that smaller models fail to learn. While this advantage is often assumed to stem from large models representing more structure within the training data, our results identify a complementary benefit: scale can also make learning more robust to adverse effects during training. We define and examine one such effect, primacy bias, as the persistent impairment in fitting late data due to exposure to early data distributions. Under mechanistic analysis of MLPs, we find that overparameterized models are able to circumvent parameters heavily recruited to fit early data distributions, which may be difficult to repurpose due to loss of plasticity.

Primacy bias becomes particularly consequential under sequential data order constraints in pretraining. In these settings, foundation models often encounter heterogeneous data distributions sequentially rather than jointly, which can favor fitting early distributions over late ones. When late data represents desirable qualities such as skills in coding, tool-use, and reasoning, this can be especially harmful. In sequential pretraining up to the billion-parameter scale, we find that primacy bias reliably impairs models over several types of data distributions. To help allocate learning capacity more efficiently across all data distributions, we introduce Exposure Therapy, a regularization that introduces a subset of late data early in pretraining where models may exhibit heightened plasticity. While adding negligible extra computational cost, this provides a cheaper alternative to scaling models, and demonstrates how some benefits of model scale can be recovered algorithmically.

\section*{Reproducibility Statement}
We provide the code to toy vision experiments along with their mechanistic analysis, foundation model experiments, and Exposure Therapy, and additionally outlining hyperparameters and most experimental details for pretraining experiments within the Appendix. All datasets used are publicly available, and we include scripts to retrieve them.

\clearpage

\bibliography{references}
\bibliographystyle{plainnat}

\appendix
\newpage

\section{Primacy Sensitivity in Toy Experiments}
\label{app:toy_setup}

We provide additional implementation details for the toy experiments in Section~\ref{sec:controlled}.
All experiments compare paired \emph{target-first} and \emph{OOD-first} runs at the
same model scale and random seed. The two runs use the same architecture,
initialization distribution, optimizer, and total number of updates, and differ only
in the objective optimized during Phase~1. No parameters, optimizer state, or
learning-rate state are reset at the phase boundary.

\paragraph{Shared two-phase protocol.}
For the toy experiments, the OOD objective is implemented using real inputs from the corresponding training distribution paired with fixed random targets. Thus, the corruption is in the input--target mapping rather than in the input marginal.

For each run, we construct a fixed Phase-1 corruption set $A$. In the OOD-first
condition, all Phase-1 updates are sampled from $A$; in the target-first condition,
all Phase-1 updates are sampled from the target task. During Phase~2, both
conditions use the same target-dominated schedule: every fourth optimization step
is an OOD update and the remaining three are target updates. Phase-2 OOD updates
use a separately sampled fixed corruption set $B$ with newly generated random
targets. Set $A$ itself is therefore never re-presented during Phase~2. Training
examples are sampled with replacement within each source.

\paragraph{Vision tasks.}
We evaluate two-hidden-layer MLPs on MNIST, Fashion-MNIST, and KMNIST.  Vision models are trained for 30,000 Phase-1 updates followed by 120,000 Phase-2 updates, which we find is appropriate for convergence. We use separate output heads for the target and OOD objectives while sharing the hidden representation, so that any cross-phase interference reflects structure learned in the shared features rather than incidental alignment or misalignment between target class identities.

\paragraph{Optimization and evaluation.}
All toy models are optimized with Adam and cross-entropy loss using a constant learning rate.

\section{OOD-Set Accuracy Across the Full Parameter Ladder}
\label{app:full-corruption-retention}

\begin{figure}[h]
    \centering
    \includegraphics[width=\linewidth]
    {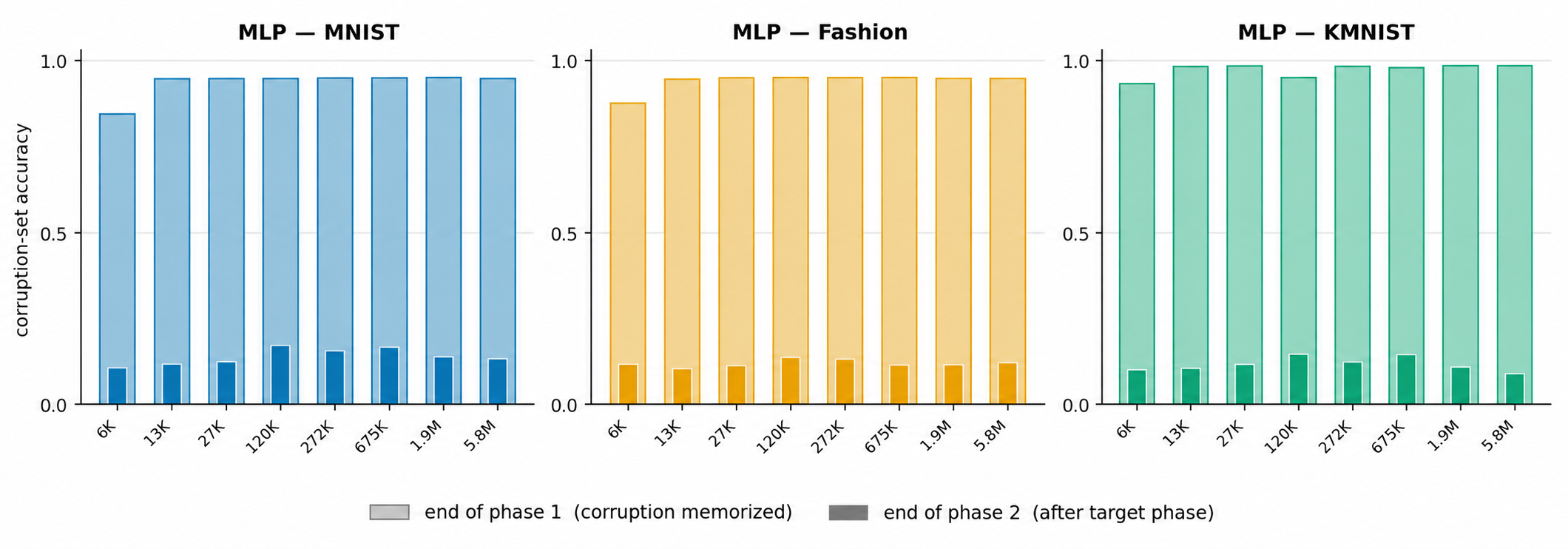}
    \caption{
        \textbf{The early OOD mapping is no longer behaviorally expressed after target training across the full parameter ladder.}
        Each panel shows one dataset, with model parameter
        count on the horizontal axis and accuracy on the original Phase-1 OOD-set on the vertical axis. Pale bars report accuracy at the end of Phase-1, when the OOD-first models have been trained on the fixed random mapping; dark bars report accuracy on the same
        examples after completing the target-dominated Phase-2. Small models fit the OOD data to high accuracy during Phase-1, and drop much lower after Phase-2. Models demonstrate the primacy bias effect without necessarily expressing OOD data seen initially.
    }
    \label{fig:junk-invisibility-full}
\end{figure}

We demonstrate that, while training on OOD data early on can impair the ability to fit different training distributions afterward, this doesn't necessarily mean that models exhibit high accuracy on early OOD data. Figure ~\ref{fig:junk-invisibility-full} demonstrates this trend.

\section{Overlap Mechanism}
\label{sec:overlap}

To understand and quantify how early learning can limit the capacity of models on downstream distributions, we compare overlap in usable neurons before and after Phase-2, where they are exposed to target distributions.

We record activation snapshots at the end of each phase using probe data from the OOD and target distributions. For a
hidden unit \(u\), we define its activation score at phase \(t\) as its mean
absolute post-activation response:
\begin{equation}
    s_t(u)
    =
    \frac{1}{|\mathcal{P}_t|}
    \sum_{x \in \mathcal{P}_t}
    \left|a_t(x;u)\right|,
    \label{eq:activation-score}
\end{equation}
where \(\mathcal{P}_t\) is the corresponding probe set and \(a_t(x;u)\) is the
activation of unit \(u\). These units are post-ReLU hidden activations in MLPs.

For each phase, we select the top fraction \(f\) of units according to
Equation~\eqref{eq:activation-score}. Let
\(A_{\mathrm{cor}}^{(f)}\) denote the units most strongly recruited at the end of
the OOD phase and \(A_{\mathrm{tgt}}^{(f)}\) the units most strongly
recruited after target training. We measure cross-phase representational overlap using the
Jaccard index:
\begin{equation}
    \operatorname{R}^{(f)}
    =
    \frac{
        \left|
            A_{\mathrm{cor}}^{(f)}
            \cap
            A_{\mathrm{tgt}}^{(f)}
        \right|
    }{
        \left|
            A_{\mathrm{cor}}^{(f)}
            \cup
            A_{\mathrm{tgt}}^{(f)}
        \right|
    }.
    \label{eq:jaccard-overlap}
\end{equation}

In MLPs, the top-$f$ selection and Jaccard overlap in Equation~\eqref{eq:jaccard-overlap} are computed independently within each hidden layer. We report $\operatorname{R}^{(f)}$ as the mean of these layer-wise Jaccard overlaps across hidden layers. Figure~\ref{fig:mechanism} plots how this overlap changes across scales. 
\section{Pretraining Experimental Details}
\label{sec:pretraining-details}

We provide the full experimental details for the sequential-domain pretraining experiments in Section~\ref{sec:foundation}. These experiments use byte-level causal Transformers trained at approximately 100M, 500M, and 1B parameters. For each model scale, we evaluate English-first and OOD-first training across two English corpora and four OOD corpora.

\paragraph{Model architecture.}
All models operate directly on bytes with a vocabulary size of 256 and a sequence length of 512. We use 7-layer causal Transformers, pre-LayerNorm blocks, and a final LayerNorm before the output head. The feed-forward activation is GeLU, and there is no dropout. Model width is scaled to obtain the three parameter counts used in our experiments:
\begin{center}
\begin{tabular}{lrr}
\toprule
Model scale & Hidden width & Parameters \\
\midrule
100M & 1088 & 100,650,048 \\
500M & 2432 & 499,545,216 \\
1B   & 3456 & 1,007,151,232 \\
\bottomrule
\end{tabular}
\end{center}

\paragraph{Training schedule.}
Training is divided into two equal-length phases. The total number of optimization steps increases with model scale:
\begin{center}
\begin{tabular}{lrrr}
\toprule
Model scale & Phase-1 steps & Phase-2 steps & Total steps \\
\midrule
100M & 9,000  & 9,000  & 18,000 \\
500M & 27,000 & 27,000 & 54,000 \\
1B   & 43,000 & 43,000 & 86,000 \\
\bottomrule
\end{tabular}
\end{center}

For each English--OOD corpus pair, the training conditions differ in the data used during Phase~1. English-first models train on the corresponding English corpus during Phase~1, whereas OOD-first models train on the corresponding OOD
corpus. Phase~2 is identical across the two conditions and consists of a fixed \(75\%/25\%\) mixture of English and OOD updates. We implement this mixture deterministically, with every fourth optimization step drawn from the OOD
corpus and the remaining three steps drawn from English. 
Thus, the distinction between English-first and OOD-first training is restricted to the first half of training.

The phase boundary does not trigger any training reset. Model parameters, optimizer state, and the learning-rate schedule continue uninterrupted from Phase~1 into Phase~2.

\paragraph{Optimization.}
We optimize all models with Adam using
\(\beta_1=0.9\) and \(\beta_2=0.999\), with no weight decay. The learning rate is \(1.5\times10^{-4}\) at all model scales. We linearly warm up the learning rate for the first 1,000 optimization steps and then hold it constant for the
remainder of training. Gradients are clipped to a maximum norm of \(0.5\).

All experiments use a batch size of 32 sequences, with 512 bytes per sequence, for a total of 16,384 byte tokens per optimization step. 

\paragraph{Pretraining corpora.}
We use CodeSearchNet for code, OpenWebMath for mathematics, and Wikipedia for German and Finnish. The English target distributions are FineWeb and TinyStories. Each OOD corpus is paired separately with each English corpus, yielding the code, mathematics, German, and Finnish conditions reported in
Section~\ref{sec:foundation}.

\paragraph{Evaluation.}
Final evaluation uses 100k bytes of random byte-sequences of withheld data, each 512 bytes in length, at a 50/50 split between each distribution seen. We report \emph{English loss} using only the held-out English data. We additionally report \emph{aggregate loss} over the combined held-out English and OOD evaluation
sets, thereby measuring performance across both distributions encountered during training. 

\section{Exposure Therapy Details}
\label{app:expose_details}

\begin{algorithm}[H]
\DontPrintSemicolon
\SetKwInOut{Input}{Input}
\SetKwInOut{Output}{Output}
\caption{Training with Exposure Therapy}
\label{alg:anchor}

\Input{
model \(f_\theta\);
ordinary training dataset \(\{\mathcal{D}_t\}_{t=1}^{T}\);
target-data pool of late training data \(\Dgood\);
exposure rate \(a\in[0,1]\);
exposure set size \(s\);
number of training steps \(T\);
batch size \(B\);
optimizer \(\operatorname{Opt}\)
}

\Output{parameters \(\theta_T\)}

\BlankLine
Fix an exposure set
\[
    \mathcal{A}\subset\Dgood,
    \qquad |\mathcal{A}|=s,
\]
which remains part of the ordinary training pool: its examples are still drawn by
ordinary updates, so Exposure Therapy adds no data and withholds none\;

\BlankLine
\For{\(t\gets1\) \KwTo \(T\)}{

    \(b_t\sim\operatorname{Bernoulli}(a)\)\;

    \eIf{\(b_t=1\)}{
        draw \(B\) from
        \(\{1,\dots,s\}\) and let \(Z_t\) be the corresponding examples of
        \(\mathcal{A}\)
        \tcp*{exposure update}
    }{
        \(Z_t\sim\operatorname{Batch}(\mathcal{D}_t,B)\)
        \tcp*{ordinary update}
    }

    \(g_t\gets
    \nabla_\theta
    \left[
        \dfrac{1}{|Z_t|}
        \sum_{z\in Z_t}
        \ell(f_\theta;z)
    \right]\)\;

    \(\theta_t\gets
    \operatorname{Opt}(\theta_{t-1},g_t)\)\;
}

\end{algorithm}

Algorithm 1 illustrates a simple training loop with Exposure Therapy. Here, \(\mathcal{D}_t\) denotes the dataset from which the unmodified training procedure would draw its batch at optimization step \(t\). The ordinary training distribution can change over the training trajectory, as in sequential pretraining. Exposure Therapy fixes an exposure set \(\mathcal{A}\subset\Dgood\) containing \(s\) distinct
examples from late training data. The exposure set is not held out: it remains part of the ordinary training pool and its examples are still drawn by ordinary updates, so Exposure
Therapy introduces no additional data. At each step, \(b_t\sim\operatorname{Bernoulli}(a)\) determines
whether the ordinary update is replaced by an exposure update. If \(b_t=0\), the model receives a batch from the scheduled ordinary dataset \(\mathcal{D}_t\). If \(b_t=1\), the model instead receives a batch of \(B\) examples drawn from \(\mathcal{A}\). Exposure updates therefore sample the exposure set independently at every step rather than traversing it in a fixed order, and because they replace ordinary updates rather than supplement them, the total number of
optimization steps is unchanged.

There are two hyperparameters introduced in ET: exposure rate and exposure set size. Conceptually, exposure set size can be seen as a heuristic to estimate \emph{the capacity that should remain available for late training}, while exposure rate reinforces \emph{how strongly training is constrained to preserve that capacity}. A larger exposure set size may help models find a more accurate estimate of how capacity should be reserved for this distribution later on. We find that smaller models that are more constrained in terms of learning capacity relative to their datasets require both a more accurate estimate of the target distribution and more frequent reinforcement of those updates, to oppose pressure that may stem from small model scale.

\section{Exposure Therapy and Model Scale Experimental Details}
\label{sec:et_scale}

Section~\ref{sec:anchor} demonstrates how exposure dosage decreases as models become more overparameterized in toy language models. We list those experimental details here.

For a model with \(N\) parameters, let
\(\operatorname{Acc}_{\mathrm{tgt}}(N)\) denote the final target accuracy under
target-first training, and
\(\operatorname{Acc}_{\mathrm{ET}}(a,c;N)\) the accuracy obtained with Exposure Therapy.
We define the relative target-first performance as
\begin{equation}
    M_{\mathrm{ET}}(a,c;N)
    =
    \frac{
        \operatorname{Acc}_{\mathrm{ET}}(a,c;N)
    }{
        \operatorname{Acc}_{\mathrm{tgt}}(N)
    }.
    \label{eq:anchor-relative-performance}
\end{equation}
Thus, \(M_{\mathrm{ET}}=0.75\) means that the ET model reaches \(75\%\) of
the target-first model's accuracy, \(M_{\mathrm{ET}}=1\) means that it
matches the target-first model, and \(M_{\mathrm{ET}}>1\) means that it
outperforms the target-first reference.

We evaluate the minimal ET dose required to reach 90\% of the target-first
model's performance.
\begin{equation}
    a_{0.9}^\star(c;N)
    =
    \min
    \left\{
        a \in \mathcal{G}_a :
        \operatorname{Acc}_{\mathrm{ET}}(a,c;N)
        \geq
        0.9\,\operatorname{Acc}_{\mathrm{tgt}}(N)
    \right\},
    \qquad
    \label{eq:required-anchor-rate}
\end{equation}
where \(\mathcal{G}_a\) denotes the tested grid of exposure dosage.
Equivalently,
\(a_{0.9}^\star(c;N)\) is the smallest tested dosage satisfying
\(M_{\mathrm{ET}}(a,c;N)\geq 0.9\).

Figure~\ref{fig:anchor-threshold} plots how this minimum required dose changes across model scales.

\section{Exposure Therapy Pretraining at 100M and 500M Scales}
\label{app:more_scales}

\begin{figure}[h]
    \centering
    \includegraphics[width=\linewidth]{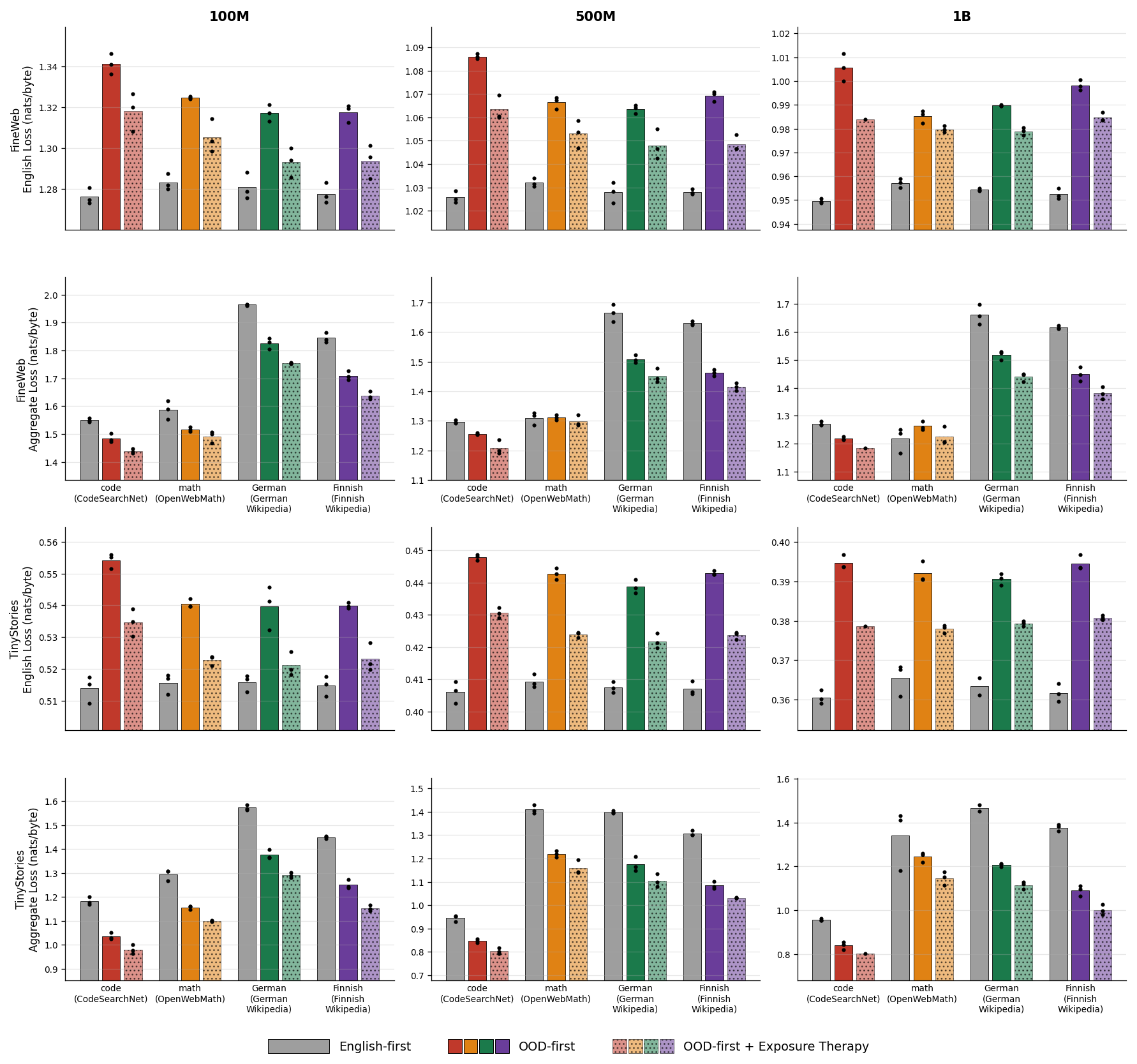}
    \caption{
        \textbf{Exposure Therapy improves sequential pretraining across model scales.}
        Columns show 100M, 500M, and 1B parameter models. The first two rows report, respectively, held-out English loss and aggregate loss when
        FineWeb is the English target distribution; the bottom two rows report the corresponding quantities for TinyStories. Within each OOD-corpus
        group, gray bars denote English-first training, solid colored bars denote OOD-first training, and dotted colored bars denote OOD-first training
        with a \(30\%\) Exposure Therapy rate and 10\% exposure set size. Black points denote individual
        runs. Lower values are better. Across all three model scales and both English target distributions, OOD-first training increases subsequent English loss relative to the English-first reference, while Exposure Therapy consistently reduces this impairment. Exposure Therapy also
        reduces aggregate loss relative to unregularized OOD-first training across the tested conditions.
    }
    \label{fig:foundation-et-all-scales}
\end{figure}

We extend the Exposure Therapy pretraining experiments of
Section~\ref{sec:foundation-results} to the 100M and 500M parameter scales, and include the 1B results for comparison. We use the same sequential-domain pretraining setup across scales: models are trained either English-first,
OOD-first, or OOD-first with a \(30\%\) Exposure Therapy rate and 10\% exposure set size, with code,
mathematics, German, and Finnish as the OOD distributions. We evaluate these on both English-specific test loss and aggregate test loss on all data, using the same metrics as Figure~\ref{fig:foundation-et}. Figure~\ref{fig:foundation-et-all-scales} demonstrates that this effect is consistent across all model scales.

\section{Ablation Experiments Under Varying Pretraining Conditions}
\label{sec:pretraining_ablation}

Our pretraining experiments in Section~\ref{sec:foundation} and Section~\ref{sec:anchor} assume several aspects of pretraining that may not extend to all pretraining setups. To indicate the broader applicability of Exposure Therapy and this work, we discuss those design choices in this section and perform ablation experiments without these assumptions. 

\subsection{Reverse Order Pretraining}

\begin{figure}[t]
    \centering
    \includegraphics[width=\linewidth]{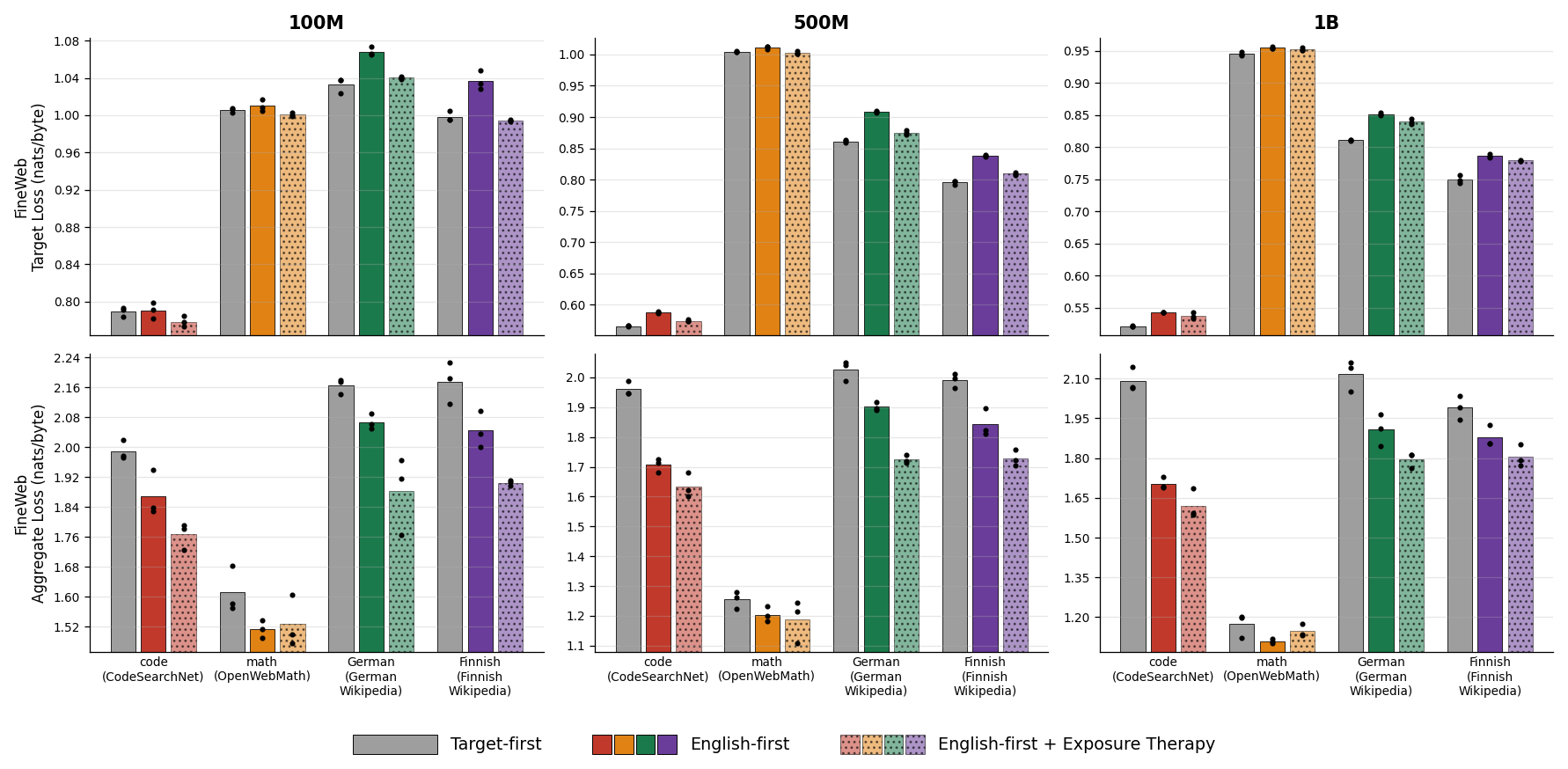}
    \caption{
        \textbf{Exposure Therapy remains effective under reverse-order pretraining.}
        Columns show 100M, 500M, and 1B parameter models. The top row reports held-out code/math/German/Finnish target loss, while the bottom row reports aggregate loss over FineWeb and the paired code, mathematics, German, or Finnish distribution; lower is better. Within each distribution, gray bars denote target-first training, solid colored bars denote English-first training, and dotted bars denote English-first training with Exposure Therapy. Black points denote individual runs. Across model scales, ET generally improves late-distribution learning relative to unregularized English-first training and consistently improves aggregate performance, demonstrating that its benefits persist when broad English data is encountered before more specialized distributions.
    }
    \label{fig:reverse_pretraining}
\end{figure}

Our pretraining experiments measure the impairment to English learning by seeing other distributions such as code, math, or other languages early in training. While this demonstrates the central claim of this work (early OOD data can impair later learning), most pretraining is done in the reverse order. Broad English natural language data is typically seen first, with distributions like code, math, and other languages seen second. We perform ablation experiments across the 100M, 500M, and 1B parameter models to demonstrate that this effect is persistent, and that Exposure Therapy improves both late learning on specialized distributions and overall loss across all distributions in these settings. Figure~\ref{fig:reverse_pretraining} demonstrates this effect.

\subsection{Persistent Effects to Retain Early Structure}
Frontier foundation models can extend beyond their training distribution in ways that other models cannot. Typically, this is associated with being able to reconcile task and data-specific qualities during training, converging to a more optimal solution across all training data rather than one solution or the other. For this reason, most pretraining pipelines train foundation models (and LLMs) over a broad range of natural language data before they specialize into more specific distributions of data, because general semantic and grammatical rules are maintained and even central to learning later distributions and behavior, like reasoning.

Because our experiments don't train foundation models within a comparable margin to frontier models (or even to a conversational level), training naively over two distributions (even at the billion-parameter scale) can be insufficient to retain early learned structures when trained on new distributions, due to a poor capability to reconcile data-specific differences across distributions in small experiments. To approximate this effect, we mix a minority of OOD data into late learning in our experiments through a deterministic mixture, with every fourth optimization step drawn from the OOD corpus and the remaining three steps drawn from the target distribution. This provides a weak training signal to maintain some degree of earlier structure, and is sufficient to cause models to fail to learn late data even when they have sufficient capacity and it is in the majority.

\begin{figure}[t]
    \centering

    \begin{minipage}[t]{0.49\linewidth}
        \centering
        \includegraphics[width=\linewidth]{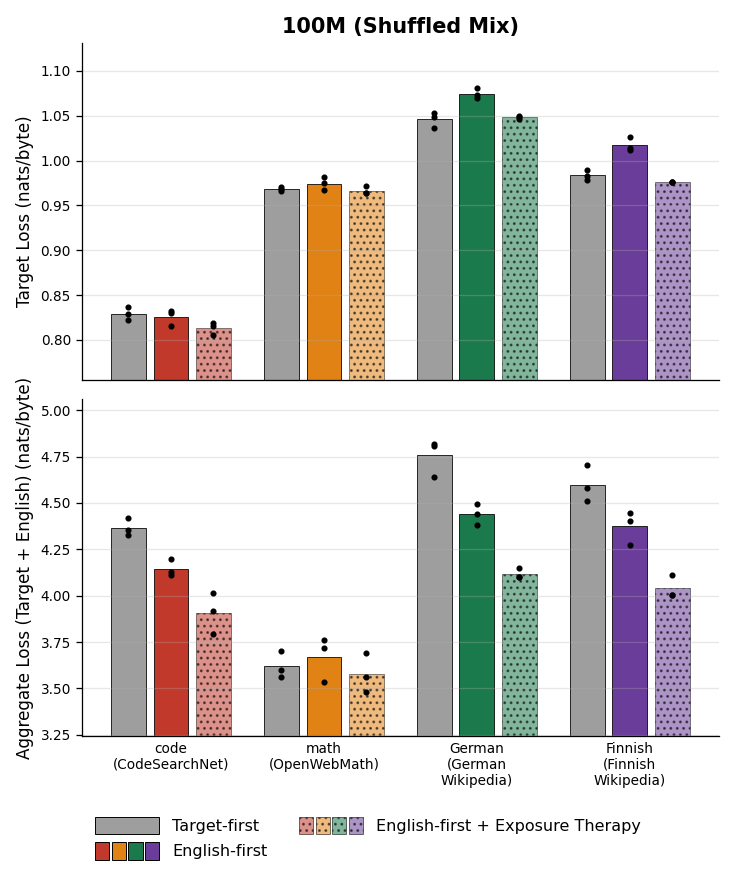}
        \vspace{-0.4em}

        \textbf{(a) Shuffled Phase-2 mixture}
    \end{minipage}
    \hfill
    \begin{minipage}[t]{0.49\linewidth}
        \centering
        \includegraphics[width=\linewidth]{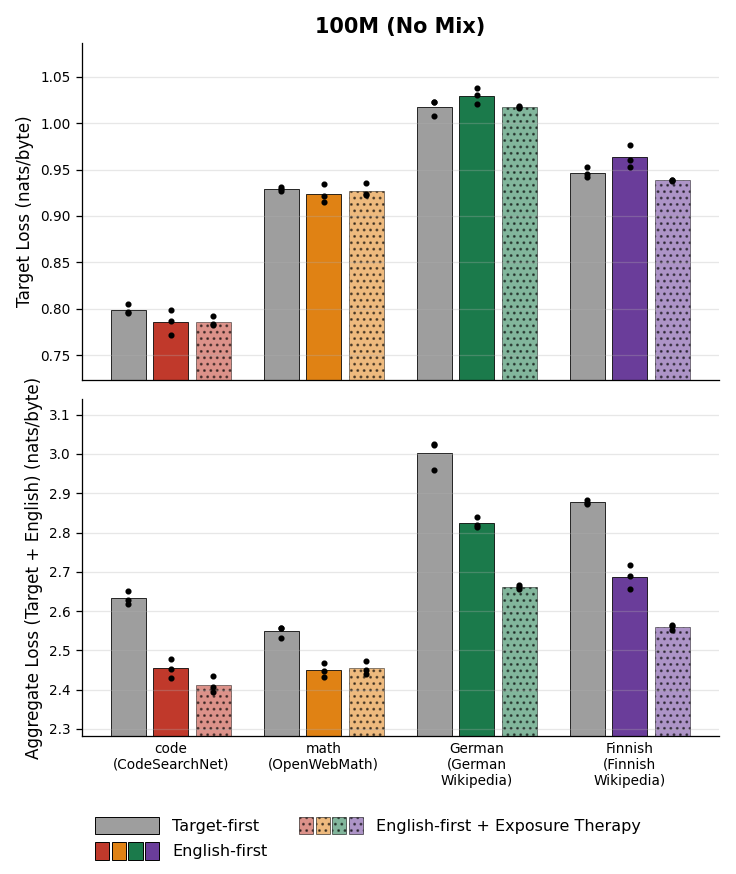}
        \vspace{-0.4em}

        \textbf{(b) No Phase-2 mixture}
    \end{minipage}

    \caption{
        \textbf{Exposure Therapy under alternative Phase-2 training conditions.}
        Results are shown for 100M-parameter models across code, mathematics, German, and Finnish target distributions. Gray bars denote target-first training, solid colored bars denote English-first training, and dotted bars denote English-first training with Exposure Therapy. Black points show individual runs. Top panels report held-out target loss and bottom panels report aggregate loss over the target and English distributions; lower is better.
        \textbf{(a)} Phase~2 uses the same target-dominated mixture as the main experiments, but English and target updates are randomly shuffled rather than deterministically interleaved. Exposure Therapy reduces both target and aggregate loss across the tested distributions.
        \textbf{(b)} Phase~2 contains only the target distribution, removing continued English updates. Exposure Therapy produces little change in target loss for code and mathematics, but improves target loss for German and Finnish and reduces aggregate loss across all four distributions.
    }
    \label{fig:persistent-structure-ablations}
\end{figure}

We relax this assumption through two ablation experiments in reverse order pretraining, which can better match practical methods. First, we train using random shuffling over the late learning (Phase-2) data between the target and OOD data, rather than training it deterministically, which can more closely mimic how signals to retain earlier structure are more inconsistent in pretraining. Figure~\ref{fig:persistent-structure-ablations} (left) demonstrates that these results are consistent with the deterministic mixture.

Next, we examine how primacy bias changes with no mixture of OOD data in late learning at all. Figure~\ref{fig:persistent-structure-ablations} (right) demonstrates these results. Without the presence of OOD data in late learning, distributions like code and math that are similar to English often see negligible increases in performance with Exposure Therapy. On distributions that have far less reusable structure (other languages, like German and Finnish), Exposure Therapy still improves late learning. However, aggregate loss over all distributions improves more consistently over all data types. Without the presence of OOD data in late learning, some models can often catastrophically forget the first distribution to better fit later distributions. Interestingly, Exposure Therapy's encouragement to reconcile data-specific differences seems to improve how \textit{previously learned structure is retained}, reducing catastrophic forgetting in these settings, in addition to improving late learning. Because this work focuses on improving late learning specifically, we leave further analysis of how ET can reduce catastrophic forgetting to future work, and include these results solely to demonstrate broader robustness of Exposure Therapy under different pretraining conditions.

\end{document}